\documentclass[final]{cvpr}
\usepackage{subfigure}
\usepackage{times}
\usepackage{epsfig}
\usepackage{graphicx}
\usepackage{amsmath}
\usepackage{amssymb}
\usepackage{multirow}
\usepackage{comment}
\usepackage{appendix}

\usepackage[pagebackref=true,breaklinks=true,colorlinks,bookmarks=false]{hyperref}

\def\cvprPaperID{5941} 
\def\confYear{CVPR 2021}

\begin{document}

\title{Glance and Gaze: Inferring Action-aware Points for One-Stage Human-Object Interaction Detection}

\author{Xubin Zhong$^1$ \quad Xian Qu$^{1}$
 \quad Changxing Ding$^{1,2}$\thanks{Corresponding author. } \quad Dacheng Tao$^3$\\
$^1$ South China University of Technology  \quad $^2$ Pazhou Lab, Guangzhou \quad $^3$ The University of Sydney\\
{\tt\small \{eexubin, eequxian.scut\}@mail.scut.edu.cn, chxding@scut.edu.cn, dacheng.tao@sydney.edu.au}
}

\maketitle


\begin{abstract}
Modern human-object interaction (HOI) detection approaches can be divided into one-stage methods and two-stage ones.
One-stage models are more efficient due to their straightforward architectures, but the two-stage models are still advantageous in accuracy.
Existing one-stage models usually begin by detecting predefined interaction areas or points, and then attend to these areas only for interaction prediction; therefore, they lack reasoning steps that dynamically search for discriminative cues.
In this paper, we propose a novel one-stage method, namely Glance and Gaze Network (GGNet), which adaptively models a set of action-aware points (ActPoints) via glance and gaze steps.
The glance step quickly determines whether each pixel in the feature maps is an interaction point.
The gaze step leverages feature maps produced by the glance step to adaptively infer ActPoints around each pixel in a progressive manner.
Features of the refined ActPoints are aggregated for interaction prediction. Moreover, we design an action-aware approach that effectively matches each detected interaction with its associated human-object pair, along with a novel hard negative attentive loss to improve the optimization of GGNet.
All the above operations are conducted simultaneously and efficiently for all pixels in the feature maps.
Finally, GGNet outperforms state-of-the-art methods by significant margins on both V-COCO and HICO-DET benchmarks. Code of GGNet is available at \texttt{\url{https://github.com/SherlockHolmes221/GGNet}}.
\end{abstract}

\begin{figure}[t]
	\begin{center}
		\includegraphics[width=0.48\textwidth]{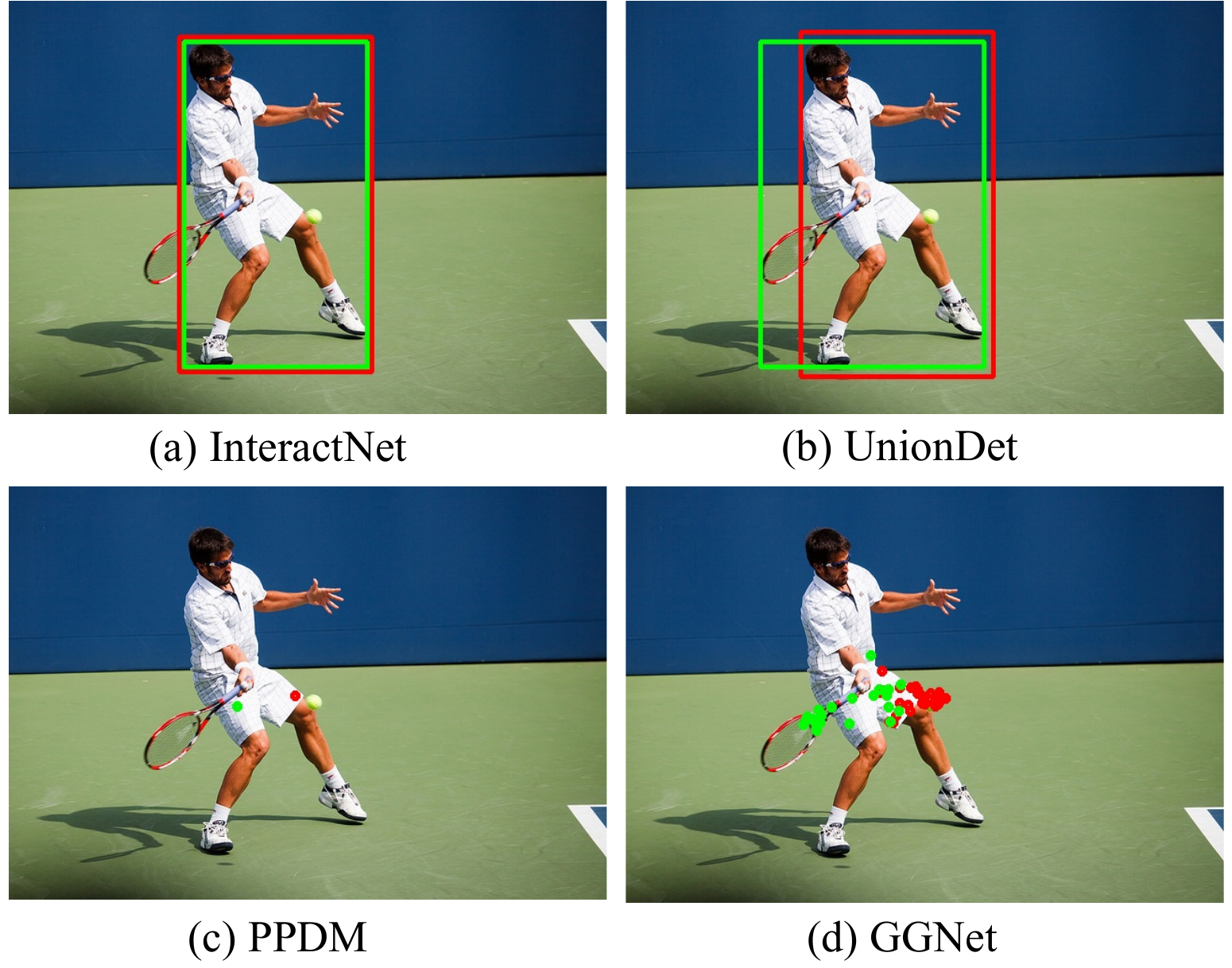}
	\end{center}
	\caption{Comparisons of interaction area definition. Green boxes or points represent the interaction area for ``hold tennis\_racket'', while the red ones stand for ``hit sports\_ball''. (a) InteractNet \cite{Gkioxari2017Detecting} uses the same human bounding box to represent the interaction area for all interactions pertaining to the person. (b) UnionDet \cite{kim2020UnionDet}  adopts the union box of one human-object pair to represent their interaction area. (c)  PPDM \cite{liao2020ppdm}  leverages the middle point of one human-object pair to represent their interaction area. (d) GGNet employs a single set of dynamic points to adaptively capture informative areas for the interaction between each human-object pair.}
	\label{Figure:cmp}
\end{figure}

\section{Introduction}
\label{sec:introduction}
Human-Object Interaction (HOI) detection is one of the fundamental tasks in human-centric scene understanding.
It involves not only detecting persons and objects in an image, but also the interactions (verbs) between each human-object pair.
The output of an HOI detection model can be represented as a set of triplets in the form of $<$$human$ $interaction$ $object$$>$, and each triplet is also referred to as a single HOI category.
For example, there are two HOI categories in Figure  \ref{Figure:cmp}, i.e. $<$$human$ $hold$ $tennis$$\_$$racket$$>$ and $<$$human$ $hit$ $sports$$\_$$ball$$>$.

According to the order in which object detection and interaction detection is performed, modern HOI detection methods can be divided into one-stage and two-stage approaches.
Two-stage methods must perform the object detection first and then identify the interactions between each possible human-object pair.
However, because the two stages are separated in this approach, these methods are usually inefficient.
By contrast, one-stage methods can perform object detection and interaction detection in parallel by first defining interaction areas (e.g. a union box or a single point).
Generally speaking, one-stage methods tend to be more efficient and structurally elegant, but two-stage methods are more accurate at present.

One of the key issues for one-stage methods is the way to represent the ``interaction area'' for each human-object pair  \cite{kim2020UnionDet}.
Existing approaches usually define this area artificially and the interaction area often face the semantic ambiguity problem.
For example, as shown in Figure \ref{Figure:cmp}(a), InteractNet  \cite{Gkioxari2017Detecting}  utilizes a human bounding box to represent the area of all interactions involving the person,
meaning that object-specific information is ignored.
UnionDet \cite{kim2020UnionDet} addresses this problem by utilizing the union box for each human-object pair as their interaction region.
However, union boxes may overlap significantly with each other (Figure \ref{Figure:cmp}(b)), which introduces ambiguity between pairs.
For its part, PPDM  \cite{liao2020ppdm} utilizes the middle point of each human-object pair as the interaction point (Figure \ref{Figure:cmp}(c)).
Although interaction points are less likely to overlap with each other under this approach, a single interaction point is often vague to represent complex interactions between a human-object pair.

With the predefined interaction areas discussed above, existing one-stage methods usually attend to the interaction area only once to predict the interaction categories. Recent works \cite{deubel1996saccade}, \cite{lan2020saccadenet} have revealed that the eyes of human beings usually move around an object to discover more cues regarding its location. Similarly, when it comes to HOI detection, people often first {\bf glance} at the scene to identify possible human-object pairs with any interaction;
they then search for cues around each pair, and finally {\bf gaze} at discriminative areas to identify the interaction class.

Accordingly, inspired by the above observation, we herein propose a novel model, named Glance and Gaze Network (GGNet), which adaptively infers a set of action-aware points (ActPoints) to represent the interaction area (Figure \ref{Figure:cmp}(d)).
 GGNet mimics the two steps taken by humans to identify human-object interactions: Glance and Gaze.
 First, GGNet quickly determines whether each pixel in the feature maps is an interaction point;
 we call it the glance step.
 Based on the feature maps in the glance step, the subsequent gaze step searches for a set of ActPoints around each pixel.
 This step then progressively proceeds to refine the location of these ActPoints.
 In brief, this step comprises two sub-steps, in which the coarse location and location residuals of ActPoints are inferred, respectively.
 Finally, GGNet aggregates features of the refined ActPoints to predict interaction categories at the interaction points.

We further propose an action-aware point matching (APM) approach designed to match each interaction with its associated human-object pair.
This matching process specifies the location of both the human and object instances for each interaction.
Existing interaction point-based methods  tend to employ a single location regressor shared by all interaction categories \cite{liao2020ppdm}, \cite{wang2020learning};
however, we observe that the interaction category affects the spatial layout of one human-object pair.
We accordingly propose to assign each interaction category a unique location regressor, which is proven in the experimentation section to be a more effective approach.

Finally, we propose a novel focal loss, namely Hard Negative Attentive (HNA) loss, to further promote the performance of GGNet.
As there are massive numbers of negative samples for each interaction classifier of the interaction point-based methods \cite{liao2020ppdm}, \cite{wang2020learning}, a serious imbalance problem exists between the positive and negative samples for each interaction category.
We thus develop an efficient approach to address this problem by inferring and highlighting hard negative samples.
Hard negatives are inferred between meaningful HOI categories containing  the same object.
For example, we can infer a hard negative sample ``repair bicycle'' according to the labeled positive sample ``carry bicycle'', unless ``repair bicycle'' is labeled as positive;
in this way, the decision boundary between easily confused interaction categories can be clarified.

Both APM and HNA loss can be readily applied to other interaction point-based HOI detection methods. We conduct extensive experiments on the two most popular HOI detection databases, i.e. V-COCO \cite{gupta2015visual} and HICO-DET \cite{chao2018learning}. Experimental results demonstrate that our proposed GGNet consistently outperforms start-of-the-art methods.
\begin{figure*}[t]
	\begin{center}
		\includegraphics[width=0.8\textwidth]{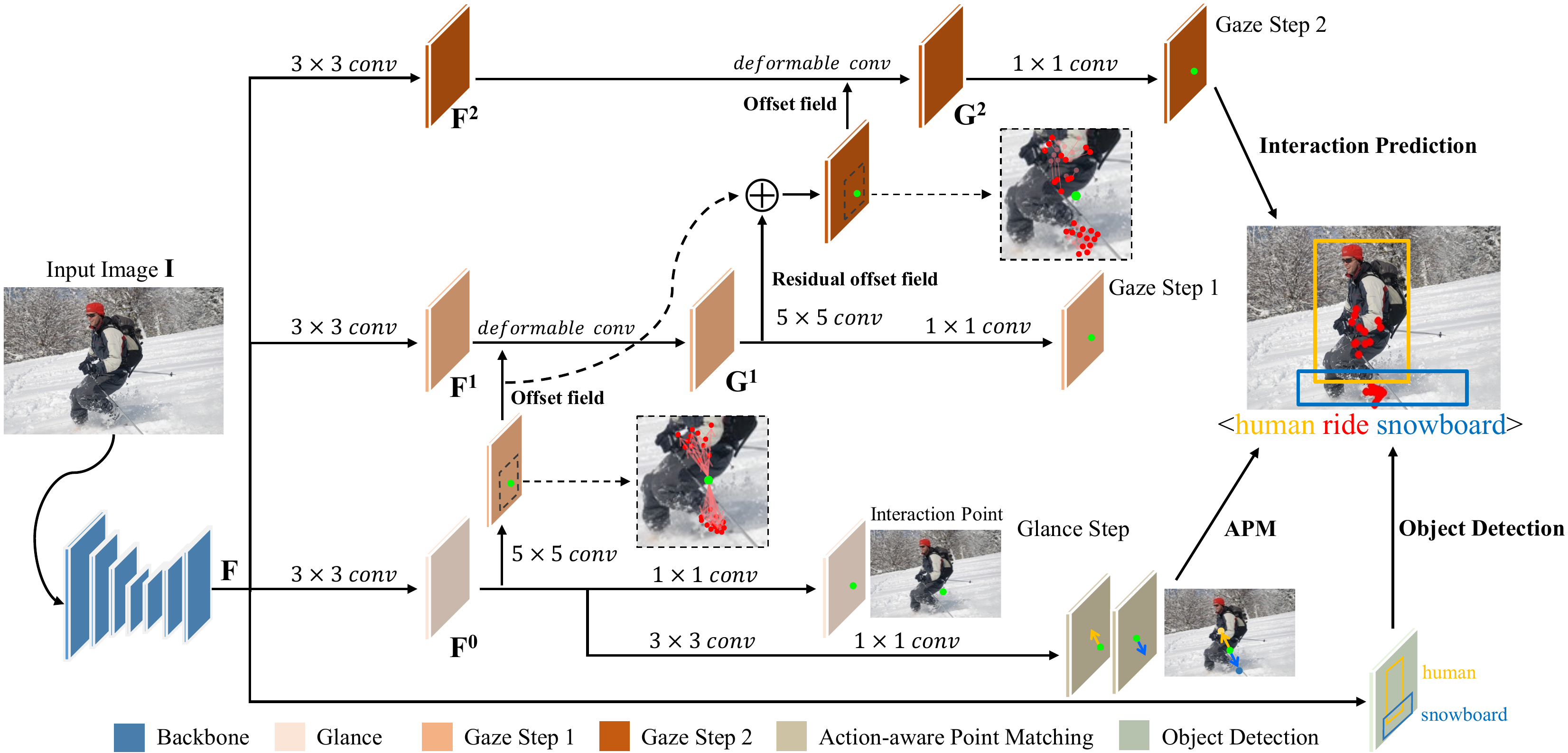}
	\end{center}
	\caption{ Overview of GGNet in the training stage. GGNet includes three main tasks, namely interaction prediction, human-object pair matching, and object detection. The three tasks share the same backbone model. The interaction prediction task includes one Glance step and two Gaze steps. The two Gaze steps infer a set of ActPoints for each pixel in feature maps. By aggregating their features into the interaction point, the second Gaze step is able to predict interactions more robustly. The human-object pair matching task is realized by the Action-aware Point Matching (APM) module, which bridges the interaction prediction and object detection tasks. In the testing stage, the glance step and the first gaze step are only utilized to infer ActPoints, while the other layers of the two steps are removed. $\oplus$ denotes the element-wise addition operation. Best viewed in color. }
	\label{Figure:overview}
\end{figure*}

\section{Related Works}
\label{sec:related_work}
\noindent{\bf Two-Stage Methods.}
Most existing HOI detection methods belong to the two-stage category.
These methods typically perform object detection first, then pair the human and object proposals for interaction recognition.
Various types of features are utilized to promote the  detection accuracy:
for example, human-object spatial feature encoded using their bounding box locations are widely adopted in two-stage methods
\cite{gupta2019no}, \cite{li2019transferable}, \cite{wan2019pose}, \cite{ulutan2020vsgnet}, \cite{lin2020gps}, \cite{wang2019deep}, \cite{Gao2018iCAN}.
In line with this, Liu \etal~\cite{liuamplifying} introduced a mechanism that encodes the fine-grained spatial layout of human-object pairs.
The spatial layout is obtained using human parsing and object segmentation tools.
Wan \etal~\cite{wan2019pose} and Zhou \etal~\cite{zhou2019relation} utilized key points on the human body to crop human part features. Different model structures have also been developed to promote HOI detection performance. For example, several recent works have utilized graph convolutional networks to integrate human and object appearance features \cite{ulutan2020vsgnet}, \cite{gao2020drg}, \cite{xu2019learning}, \cite{Qi2018Learning}.
Moreover, Gao \etal~\cite{Gao2018iCAN} and Wang \etal~\cite{wang2019deep} made use of attention mechanisms to capture context information.
Other works have also investigated the semantic meaning of verbs and the relationships between them:
for example, Zhong \etal~\cite{zhong2020polysemy, zhong2021polysemy}  promoted HOI detection accuracy by overcoming the so-called verb polysemy problem,
while Kim \etal~\cite{kim2020detecting} modeled the relationship between interaction categories by means of action co-occurrence matrices, which were then utilized to promote HOI detection performance.

Although two-stage methods are flexible to include diverse features, they divide object detection and interaction prediction into two sequential steps, which is typically very time-consuming.

\noindent{\bf One-Stage Methods.}
Some early HOI detection works \cite{Shen2018Scaling}, \cite{Gkioxari2017Detecting} devised end-to-end models based on the Faster R-CNN object detector \cite{ren2015faster}.
Although these methods are more efficient than most two-stage methods, they adopt the same human appearance features for predicting different interactions, meaning that object-specific information in different human-object pairs is ignored.
Recent works, moreover, have proposed to perform object detection and interaction prediction in parallel branches.
For example, Liao et al. \cite{liao2020ppdm} and Wang et al. \cite{wang2020learning} defined the middle point of each human-object pair as their interaction point,
which is capable of roughly capturing both human and object appearance features.
These authors then detect interaction points in the interaction prediction branch.
Kim \etal~\cite{kim2020UnionDet} represented the interaction location of one human-object pair using the union box, then detect union boxes using networks that are similar to anchor-based object detection models \cite{ren2015faster}, \cite{lin2017feature}, \cite{lin2017focal}.
These recent one-stage methods can substantially  improve HOI detection efficiency;
however, as illustrated in Figure \ref{Figure:cmp}, their definitions of the interaction areas or points remain relatively coarse and may thus introduce  ambiguity into HOI detection.

In this paper, to handle the semantic ambiguity problem associated with interaction areas, we propose to represent these area using a set of dynamic ActPoints.
These ActPoints are adaptively inferred around each interaction point, after which their features are aggregated to improve  the interaction recognition accuracy.

\section{Methods}
\label{sec:method}
\subsection{Overview}
The architecture of GGNet is illustrated in Figure \ref{Figure:overview}.
Similar to PPDM \cite{liao2020ppdm}, GGNet breaks HOI detection into three main tasks, namely interaction prediction, human-object pair matching, and object detection.
These three tasks all share the same backbone.
The second task bridges the first and the third tasks via associating each detected interaction with a single human-object pair. The object detection task is realized according to \cite{liao2020ppdm}.
GGNet improves the interaction prediction task with a novel glance-and-gaze strategy.
It also promotes the accuracy of the second task with an Action-aware Point Matching (APM) module.

\subsection{Glance and Gaze Network}
As illustrated in Figure \ref{Figure:overview}, given an input image $ \mathbf{I}\in R^{H \times W \times 3}$, the output feature maps of the backbone can be expressed as $\mathbf{F} \in R^{\frac{H}{d} \times \frac{W}{d}\times C} $, where $d$ denotes the output stride of backbone and  $C$ denotes the number of channels.

In this paper, we adopt the same definition of interaction point as \cite{liao2020ppdm},
\cite{wang2020learning}, which is also illustrated in Figure \ref{Figure:cmp}(c).
GGNet handles the semantic ambiguity problem of the single interaction point by further inferring a set of action-aware points (ActPoints).
ActPoints adaptively capture more contextual information, and their features are aggregated to predict the interaction category.
The set of ActPoints around one interaction point can be represented as:
\begin{equation}
    \mathcal{P} = \{(x_{k}, y_{k})\}^{n}_{k=1},
    \label{eq:ps}
\end{equation}
where $n$ is the total number of ActPoints sampled for an interaction. We empirically set $n$ as 25. The location of ActPoints are inferred progressively by mimicking human’s visual system \cite{deubel1996saccade} with one glance step and two gaze steps.

\noindent{\bf Glance Step.}
In this step, GGNet quickly determines whether each pixel in $\mathbf{F} $ is an interaction point.
As shown in Figure \ref{Figure:overview}, this step is realized using  a 3 $\times$ 3 convolutional (Conv) layer with ReLU, followed by a 1 $\times$ 1 Conv layer and a sigmoid layer.
The size of heatmaps produced by the sigmoid layer is $\frac{H}{d} \times \frac{W}{d}\times V $, where $V$ denotes the number of interaction categories.
We apply a $V$-dimensional element-wise focal loss  \cite{liao2020ppdm} to the heatmaps as supervision for the inference of interaction categories.
Due to this supervision, the feature maps output by the 3 $\times$ 3 Conv layer, i.e. $\mathbf{F^{0}}$ in Figure \ref{Figure:overview}, are action-aware.

\noindent{\bf  Gaze Step.} This step infers the location of ActPoints via two sub-steps, which are referred to as Gaze Step 1 and Gaze Step 2, respectively.
In Gaze Step 1, $\mathbf{F}^{0}$ is employed to predict the coarse location of $n$ ActPoints for each pixel, since features in $\mathbf{F}^{0}$ have been action-aware.
Moreover, as the discriminative power of each ActPoint varies with respect to the target interaction, GGNet also predicts a weight for each  ActPoint.
Both location and weight prediction are achieved by a 5 $\times$ 5 Conv layer.
Next, we aggregate features of ActPoints as well as their weights using one deformable Conv layer \cite{zhu2019deformable}. The 5$\times$5 offset field is determined by the number of ActPoints, which are also verified in Table \ref{tab:gaze}.
To ensure that the predicted ActPoints are reasonable, the feature maps produced by this sub-step are also used for interaction prediction with the $V$-dimensional element-wise focal loss as supervision.

On its own, the above step cannot always  obtain precise locations of ActPoints;
this is because the above 5 $\times$ 5 Conv operation has a fixed field of view, while the location of the human and object instances in one pair can vary dramatically.
To address this problem, Gaze Step 2 is introduced to refine the location of ActPoints. In more detail, we aggregate the features in $\mathbf{F}^{1}$  of the coarse ActPoints using one deformable Conv layer, of which the output feature maps are  denoted as $\mathbf{G}^{1}$. Now, each pixel in  $\mathbf{G}^{1}$ has a larger field of view; $\mathbf{G}^{1}$ is then send to another 5 $\times$ 5 Conv layer to predict the residual offsets of ActPoints locations, along with their new weights. As shown in Figure \ref{Figure:overview}, the final position of ActPoints are obtained by summing their coarse location and residual offsets.
Finally, we  aggregate the features of the refined ActPoints as well as their weights using another deformable Conv layer to predict interaction categories for each pixel in $\mathbf{F}^{2} $. Similar to Gaze Step 1, the $V$-dimensional element-wise focal loss is adopted as supervision.

More formally, $(x^{(t)}_{k}, y^{(t)}_{k})$  is updated as follows:

\begin{equation}
   (x^{(t)}_{k}, y^{(t)}_{k}) =  ( x^{(t-1)}_{k},  y^{(t-1)}_{k}) + (\Delta x^{(t)}_{k}, \Delta y^{(t)}_{k}),
    \label{eq:ps3}
\end{equation}

\begin{equation}
   (\Delta x^{(t)}_{k}, \Delta y^{(t)}_{k}) = T^{(t)}_{offset}(\mathbf{G}^{t-1}),
    \label{eq:ps2}
\end{equation}
where $t$ stands for the $t$-th gaze step, while $(\Delta x^{(t)}_{k}, \Delta y^{(t)}_{k})$ denotes the predicted offset with respect to the $k$-th ActPoint’s location in the last step.
Moreover, we set the initial location of all $n$ ActPoints as $(0, 0)$ and set $\mathbf{G}^{0}$ as $\mathbf{F}^{0}$.
$ T^{(t)}_{offset}$ is a Conv layer, whose kernel size is equal to the square root of $n$. More details of the glance and gaze network can be confirmed from our open-source project.
\begin{figure}[t]
	\begin{center}
		\includegraphics[width=0.48\textwidth]{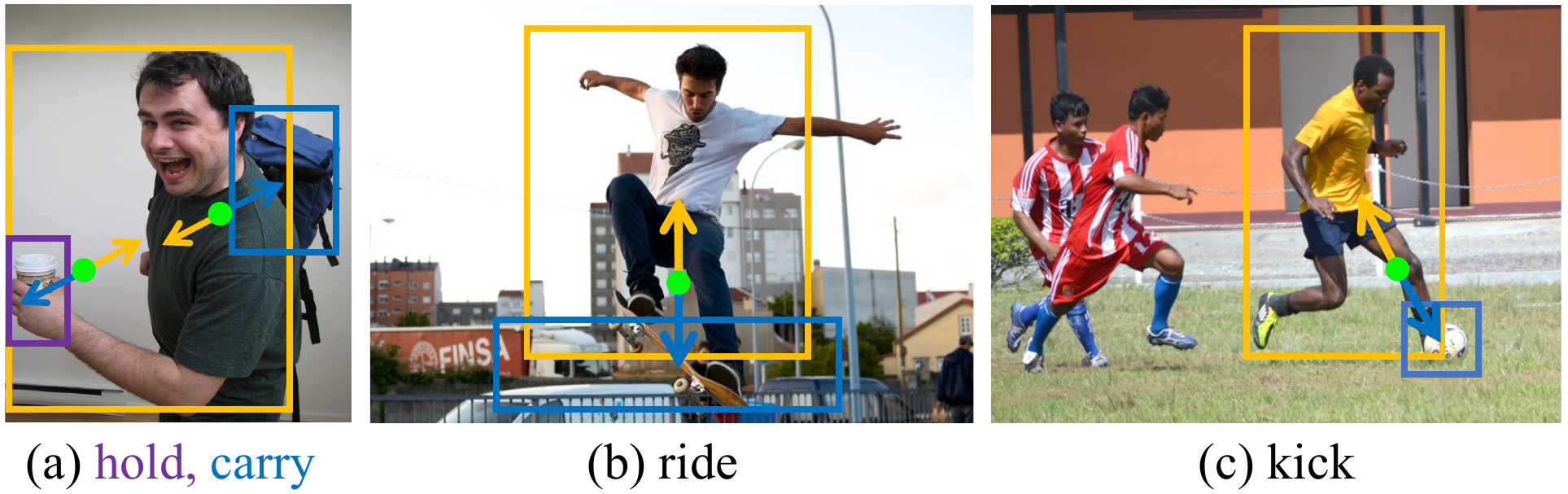}
	\end{center}
	\caption{Human-object pairs with different interaction categories present different spatial characteristics, i.e. the relative location between one human-object pair. Green points represent interaction points, each yellow arrow specifies the human instance in the pair, while each blue arrow indicates the object instance in the pair. }
	\label{Figure:APM}
\end{figure}

\noindent{\bf Action-aware Point Matching. }
To compose one HOI instance, each detected interaction point is associated with one human-object pair.
Existing works \cite{liao2020ppdm}, \cite{wang2020learning} adopt a single regressor shared by all interaction categories for this association process.
However, as shown in Figure \ref{Figure:APM}, human-object pairs with different interactions  also differ in terms of spatial characteristics.
Therefore, we propose the action-aware point matching (APM) module that assigns a unique location regressor for each interaction category.

As illustrated in Figure \ref{Figure:overview}, APM is attached to $\mathbf{F}^{0}$.
It includes one 3 $\times$ 3 and one 1 $\times$ 1 Conv layers.
The latter layer acts as regressors. Each regressor outputs a 2-dimensional offset to human point (object point) with respect to the interaction point.
Therefore, the output dimension of this layer is $\frac{H}{d} \times \frac{W}{d}\times 4V$.  In line with \cite{liao2020ppdm}, we utilize these predicted offsets to match the target human (object) proposal during inference. This process will be detailed in Section \ref{sec:TI}.

\subsection{Hard Negative Attentive Loss}
\label{sec:hn}
Recent one-stage methods \cite{liao2020ppdm}, \cite{wang2020learning} adopt the element-wise focal loss \cite{zhou2019objects}, \cite{law2018cornernet} on the output heatmaps as supervision to train their models. However, the heatmap size for each interaction category is $\frac{H}{d} \times \frac{W}{d}$, which is often a large number. Therefore, there are massive negative samples in each heatmap, which brings in the problem of imbalance between positive and negative samples. Moreover, because of the long-tailed distribution of interaction categories  \cite{chao2018learning},  \cite{gupta2015visual}, some interaction categories have very limited positive samples, which further exacerbates the imbalance problem.
Based on the above observation, we propose a novel hard negative attentive (HNA) loss to guide the model to focus more on hard negative samples for each respective interaction category.

\noindent{\bf Definition of Hard Negatives.} We infer hard negatives between meaningful HOI categories that share the same object class.
For example, we can infer a hard negative sample $<$$human$ $repair$ $bicycle$$>$ for the ``repair'' category according to  a labeled positive sample $<$$human$ $carry$ $bicycle$$>$, if $<$$human$ $repair$ $bicycle$$>$ is indeed not labeled as positive.
We do not infer hard negatives from meaningless HOI categories that have never appeared in the training set of each database, e.g.  $<$$human$ $eat$ $bicycle$$>$ and  $<$$human$ $drink$ $bicycle$$>$. The inferred sample of $<$$human$ $repair$ $bicycle$$>$ can be highlighted as a hard negative in the interaction heatmap for ``repair'' category.

\noindent{\bf Loss Formulation.}
First, we introduce the Gaussian heatmap masks $\mathbf{M} \in [-1, 1]^{\frac{H}{d} \times \frac{W}{d}\times V}$, which are used to mark both positive samples and hard negative samples.
For a ground truth HOI sample ($v_{i}$, $o_{i}$) with an interaction point located at $(x_{i}, y_{i})$, $M_{x_{i} y_{i} v_{i}}$ is set to 1 and it is also used as the center of a Gaussian distribution in the $v_i$-th channel of $\mathbf{M}$. The value  of the elements in this distribution is within $[0, 1]$.
The interaction category and object category for this HOI sample are denoted as $v_{i}$ and $o_{i}$, respectively.

Second, we infer a set of HOI samples $\{(v_{j}, o_{i})\}$ as hard negatives for the $v_j$-th interaction category, with the help of the labeled positive sample $(v_i, o_i)$.
If $(v_j,o_i)$ is not labeled as a positive sample, then we set $M_{x_{i} y_{i} v_{j}}$ as -1 and it is also used as the center of another  Gaussian distribution in the $v_j$-th channel of $\mathbf{M}$. The value of the elements in this distribution is within $[-1, 0]$.

We repeat the above two operations for each ground truth sample.
The value of remaining elements in $\mathbf{M}$ is set to 0.
Finally, the HNA loss can be represented as follows:

\begin{equation}
\mathcal{L}= -\frac{1}{N}\sum_{x y v}\left\{\begin{array}{cc}{(1-{{P}}_{x y v})^{\alpha} \log ({P}_{x y v})} & {\text { if  } {M}_{x y v}=1}, \\
{(1-{M}_{x y v})^{\beta}({P}_{x y v})^{\alpha}} \\
{\log (1- P_{x y v})} & {\text { if  } {M}_{x y v}=-1}, \\
{(1-{M}_{x y v})^{\gamma}({P}_{x y v})^{\alpha}} \\
{\log (1-{P}_{x y v})}  & {\text { otherwise},}
\end{array}\right. \\
\label{eq:hn}
\end{equation}
where $N$ is the number of ground truth interaction points in an image.
${P}_{xyv}$ is the predicted score for the interaction category $v$ at point $(x, y)$.
$\beta$ is set as 7.
$\alpha$, $\gamma$, and the parameters of the Gaussian distribution are set following \cite{liao2020ppdm}.

\subsection{Training and Inference}
\label{sec:TI}
\noindent{\bf Training.}  The overall loss function for GGNet can be represented as follows:
\vspace{-1mm}
\begin{equation}
     \mathcal{L}_{hoi} =   \mathcal{L}_{gaze2} +\lambda_{1}(\mathcal{L}_{glance} + \mathcal{L}_{gaze1} + \mathcal{L}_{m}) + \mathcal{L}_{d},
\end{equation}
where
\begin{equation}
  \mathcal{L}_{m} = \mathcal{L}_{mh} + \mathcal{L}_{mo} ,
\end{equation}
\begin{equation}
   \mathcal{L}_{d} = \mathcal{L}_{h} + \mathcal{L}_{o} + \lambda_{2}\mathcal{L}_{wh} + \mathcal{L}_{off}.
\end{equation}

$\lambda_{2}$ is set to 0.1 following [10]; $\lambda_{1}$ is set to 0.1, which is analyzed in the supplementary file.
$\mathcal{L}_{glance}$, $\mathcal{L}_{gaze1}$ and $\mathcal{L}_{gaze2}$ denote the HNA loss for the  glance step, and each of two gaze steps, respectively.
$\mathcal{L}_{mh}$ and $\mathcal{L}_{mo}$ stand for the L1 loss for matching human and object points in the APM module, respectively.
$\mathcal{L}_d$ stands for the object detection loss.
$\mathcal{L}_h$ and $\mathcal{L}_o$ are the focal loss functions to predict human and object locations.
$\mathcal{L}_{wh}$ and $\mathcal{L}_{off}$ denote the $L1$ loss for object size and center offset predictions, respectively.
$\mathcal{L}_{m}$ and $\mathcal{L}_{d}$ are realized in the same way as \cite{liao2020ppdm}.

\noindent{\bf Inference.} 
We use the output of Gaze Step 2 to obtain interaction points; while the final score is the multiplication
result of interaction prediction score from Gaze step 2 and object detection scores of the associated human-object pair. The following point matching process is the same as that in \cite{liao2020ppdm}.
First, a set of interaction $\mathcal{\hat{S}}^{i}$, human $\mathcal{\hat{S}}^{h}$ and object points $\mathcal{\hat{S}}^{o}$ are respectively selected based on their prediction confidence scores.
The number of points in $\mathcal{\hat{S}}^{i}$, $\mathcal{\hat{S}}^{h}$, and $\mathcal{\hat{S}}^{o}$ is set to 100, respectively.
Second, we associate each predicted interaction point $(\hat{x}^{i}, \hat{y}^{i}) \in \mathcal{\hat{S}}^{i}$  with a human point $(\hat{x}_{opt}^{h}, \hat{y}_{opt}^{h}) \in \mathcal{\hat{S}}^{h}$ and an object point $(\hat{x}_{opt}^{o}, \hat{y}_{opt}^{o}) \in \mathcal{\hat{S}}^{o}$  according to the predicted subject and object offsets by APM, respectively:
\begin{equation}
\label{eq:match}
\begin{aligned}(\hat{x}_{opt}^{h}, \hat{y}_{opt}^{h})=  & \underset{(\hat{x}^{h}, \hat{y}^{h}) \in \mathcal{\hat{S}}^ h}{\mathop{\arg \min} } \frac{1}{P^h_{(\hat{x}^{h},\hat{y}^{h})}} \\ &({|(\hat{x}^{i},\hat{y}^{i})-({\hat{d}^{hx}_{(\hat{x}^{i}, \hat{y}^{i})}}, {\hat{d}^{hy}_{(\hat{x}^{i}, \hat{y}^{i})}})-(\hat{x}^{h}, \hat{y}^{h})}|  ),  \end{aligned} \\
\end{equation}
where  $P^h_{(\hat{x}^{h},\hat{y}^{h})}$ denotes the object detection score for human point $(\hat{x}^{h},\hat{y}^{h})$; $(\hat{d}^{hx}_{(\hat{x}^{i}, \hat{y}^{i})}, \hat{d}^{hy}_{(\hat{x}^{i}, \hat{y}^{i})})$ denotes the predicted offset  from the interaction point to the human point by APM. The optimal object point $(\hat{x}_{opt}^{o}, \hat{y}_{opt}^{o})$ can be inferred similarly.
\section{Experimental Setup}
\label{sec:Experiments_setting}
\subsection{Datasets and Evaluation Metrics}

\textbf{V-COCO.}  V-COCO was constructed based on the MS-COCO database \cite{lin2014microsoft}. Its training and validation sets contain 5,400 images, while its testing set includes 4,946 images. It covers 80 object categories, 26 interaction categories and 234 HOI categories. The mean average precision of Scenario 1 role ($mAP_{role}$) \cite{gupta2015visual} is used for evaluation.

\textbf{HICO-DET.} HICO-DET  \cite{chao2018learning} is a large-scale HOI detection dataset with more than 150,000 annotated instances. It contains  38,118 and 9,658 images for training and testing, respectively. There are a total of 80 object categories, 117 verb categories, and 600 HOI categories. Those HOI categories with fewer than 10 training samples are referred to as ``rare'' categories and the remaining ones are called as ``non-rare'' categories; specifically, there are 138 rare and 462 non-rare categories in total. There are two modes of mAP on HICO-DET, namely the Default (DT) mode and the Known-Object (KO) mode. In DT mode, each HOI category is evaluated on all testing images; while in KO mode, one HOI is only evaluated on images that contain its associated object category.

\begin{table}[t]
	\centering
\caption{Ablation studies on each component of GGNet. DLA-34 is adopted as backbone for experiments on HICO-DET.}
	\resizebox{0.46\textwidth}{!}{
		\begin{tabular}{c cccc cc}
			\hline
			& \multicolumn{4}{c}{Components}  &\multicolumn{2}{c}{mAP}  \\
			Methods     &HNA Loss &Gaze \# 1&Gaze \# 2&APM    & V-COCO   & HICO-DET (DT) \\
			\hline
			Our Baseline    &-&-&-&-                    &51.06 &20.16 \\
			\hline
			\multirow{3}{*}{Incremental}
			&\checkmark&-&-&-              &52.51   &20.58  \\
            &\checkmark&\checkmark&-&-                  &54.03 &21.05 \\
            &\checkmark&\checkmark  &\checkmark&-             &54.43   &21.54 \\
			\hline
			\multirow{4}{*}{Drop-one-out}	&-&\checkmark&\checkmark&\checkmark &53.55 &21.03\\
			&\checkmark&-&-&\checkmark    &52.60 &21.19 \\
			&\checkmark&\checkmark&-&\checkmark          & 54.21 &21.65  \\
            &\checkmark&\checkmark&\checkmark&-            &54.43 &21.54  \\
			\hline
			\vspace{0.mm}
			GGNet   &\checkmark&\checkmark  &\checkmark&\checkmark       &\textbf{54.72} &\textbf{22.03} \\
			\hline
	   \end{tabular}}
        \label{tab:tab1}
\end{table}

\subsection{Implementation Details}
We use the Hourglass-104 model \cite{newell2016stacked}, pre-trained on the MS-COCO  \cite{lin2014microsoft},  as the backbone of GGNet.
Moreover, we adopt a lightweight network, named DLA-34 \cite{yu2018deep}, as the backbone to perform ablation studies on HICO-DET for shortening experimental cycle.  GGNet is trained using the Adam optimizer with an initial learning rate of 1.5e-5 (1.5e-4) and batch size of 7 (23) on V-COCO (HICO-DET) for 120 epochs. For all the experiments, the learning rate is reduced by multiplying 0.1 at the 90th epoch. Resolution of input images is 512 $\times$ 512 and the output stride $d$ of backbone is set to 4.
\section{Experimental Results and Discussion}
\label{sec:EXPERIMENTS}
\subsection{Ablation Studies}
We perform ablation studies on both V-COCO and HICO-DET datasets to demonstrate the effectiveness of each component of GGNet. Our baseline is constructed by removing the gaze steps from GGNet, and replacing the HNA loss and APM module with their counterparts in PPDM \cite{liao2020ppdm}.  Experimental results are tabulated in Table~\ref{tab:tab1}.

\noindent{\bf Effectiveness of the HNA Loss.}  As analyzed in Section \ref{sec:hn}, a serious problem of imbalance exists between positive and negative samples for each interaction category. We therefore propose the HNA loss to handle this problem by highlighting the hard negative samples.  The adoption of HNA loss promotes the performance of our model by 1.45\% (0.42\%) mAP on V-COCO (HICO-DET).

We further evaluate the optimal value of the  hyper-parameters  $\beta$ in in the HNA loss. The experimental results are provided in the supplementary material.

\noindent{\bf Effectiveness of the Glance-Gaze Strategy.} The glance-and-gaze strategy is developed to infer ActPoints to represent the interaction area for one human-object pair. The Gaze Step 1 is found to promote the HOI detection performance by 1.52\% and 0.47\% in terms of mAP on V-COCO and HICO-DET. Moreover, when the ActPoints is refined by Gaze Step 2, the performance is further improved by  0.40\% and 0.49\% mAP on V-COCO and HICO-DET, respectively.

\noindent{\bf Effectiveness of APM.} As the interaction category affects the human-object spatial layout in an HOI instance, we propose the APM module that assigns each interaction category  a unique location regressor to facilitate matching of both the human and object points. The adoption of APM promotes the performance of our model  by 0.29\% and 0.49\% in terms of mAP on V-COCO and HICO-DET, respectively.

\noindent{\bf Drop-one-out Study.} We further perform a drop-one-out study in which each proposed component is removed individually. In particular, as Gaze Step 2 is built based on Gaze Step 1, we remove both Gaze Steps 1 and 2 in the experiment where ``Gaze \# 1'' is dropped out.  These experimental results further demonstrate that each proposed component is indeed effective at promoting HOI detection performance.

\begin{table}[t]
       \centering
         \caption{Comparisons with variants of the Gaze Step on V-COCO.}
        \resizebox{0.26\textwidth}{!}{
        \begin{tabular}{c|c|c|c}
        \hline
        \# ActPoints     & \# Gaze Step &Sharing& $AP_{role}$    \\
        \hline
        \hline
      9      &1  &-&53.54 \\
      25 &1 &-&54.32\\
      49  &1 &-&53.81\\
       \hline
        \hline
      25    &1 &-&54.21  \\
      25      &2 &- &\textbf{54.72}\\
      25 &3 &-&54.34\\
      \hline
        \hline
      25      &2 &\checkmark &54.43 \\
\hline
\end{tabular}}
\label{tab:gaze}
\end{table}

\subsection{Comparisons with Variants of GGNet}

\noindent{\bf Comparisons with Variants of the Gaze Step.}
We compare the performance of gaze step with some possible variants by changing the number of ActPoints, the number of gaze steps, and whether layers are shared in different gaze steps. Experimental results are summarized in Table \ref{tab:gaze}.

First, we set the number of gaze steps to 1 and change the number of ActPoints. As the results show, our model achieves the best performance when the number of ActPoints is 25; this may be because a small number of ActPoints is insufficient to cover the entire interaction area, while too many ActPoints will increase the complexity of searching their locations.

Second, we compare the performance of different numbers of gaze steps. The number of ActPoints is set to 25 here. When the number of gaze steps  increases from 1 to 2,  our model is promoted by 0.51\% in terms of mAP; notably, the performance is not further promoted through the addition of more gaze steps, which may be because this increases the difficulty of model optimization.

Third, we try sharing the parameters of the two 3 $\times$ 3 Conv layers that generate $\mathbf{F^{1}}$ and $\mathbf{F^{2}}$ in Figure \ref{Figure:overview}. As Table  \ref{tab:gaze} shows, the performance of our model decreases by 0.29\% in terms of mAP. One reason for this is that Gaze Step 2 captures more fine-grained features for interaction prediction; therefore, it is better for the two gaze steps to adopt independent Conv layers.
\begin{table}[t]
       \centering
         \caption{Comparisons with variants of feature aggregation on V-COCO. ``I'', ``H'', and ``O'' denote the features of interaction point, human center point, and object center point, respectively.}
         \vspace{1mm}
        \resizebox{0.28\textwidth}{!}{
        \begin{tabular}{c|c}
        \hline
        Methods     & $AP_{role}$    \\
        \hline
        \hline
         I (our baseline)  &51.06\\
      I + H      &51.84  \\
       I + O   &51.71\\
      I  + H  +  O  &52.32 \\
       \hline
        \hline
      ActPoints w/o glance and gaze &51.66 \\
      ActPoints + glance and gaze  &\textbf{53.25} \\
        \hline
        \end{tabular}}
        \label{tab:action_aware}
\end{table}

\noindent{\bf Comparisons with Variants of Feature Aggregation}
\label{exp:fg}
 In two-stage methods, it is a common practice to aggregate features of the human instance, the object instance, and their union box for interaction prediction purposes  \cite{wan2019pose}, \cite{hou2020visual}, \cite{zhong2020polysemy}. Accordingly, in this experiment, we compare the performance of the above feature aggregation approach with our proposed glance-and-gaze strategy. The experimental results are tabulated in Table  \ref{tab:action_aware}. All methods in this table are constructed based on our baseline model. Here,  ``I + H'' (``I + O'') means that we concatenate the features of each pixel in $\mathbf{F^{0}}$ (in Figure \ref{Figure:overview}) with those of one human (object) center point. ``I + H + O'' denotes that both features of the human and object center points are aggregated.
The human and object center points are obtained via the point matching strategy in \cite{liao2020ppdm}. Moreover, ``ActPoints w/o glance and gaze'' means that we adopt one deformable Conv layer to aggregate features of the predicted coarse ActPoints on $\mathbf{F^{0}}$ for interaction prediction. The structure of these models is outlined in more detail in the supplementary material.

From the above, we can make the following observations. First, performance is promoted when either human or object features are aggregated.  However, one person may interact with different objects, while a single object may also be interacted by multiple persons.  Therefore, features for the center points of human and object instances may lack specific information for each human-object pair. By contrast, our proposed ActPoints adaptively capture features in the discriminative area for each human-object pair; therefore, it outperforms ``I + H + O'' by 0.93\% mAP.  Second, when the glance and gaze steps are omitted, the performance of ActPoints drops by 1.59\% mAP.  These comparisons demonstrate the effectiveness of our proposed glance-and-gaze strategy.
\begin{table}[t]
       \centering
       \caption{Performance comparisons on V-COCO. $^{\circ}$  denotes methods that are reproduced using their open-source codes. `A', `P', `S', and `L' represent the appearance feature, human pose feature,  spatial feature, and   language feature, respectively. }
        \vspace{1mm}
        \resizebox{0.3\textwidth}{!}{
        \begin{tabular}{c|c|c|c}
        \hline
        &Methods    & Feature   & $AP_{role}$ \\
        \hline
        \hline
        \multirow{12}*{\rotatebox{90}{Two-Stage}}
        &RPNN  \cite{zhou2019relation} & A+S+P &47.5 \\
        &TIN  (RP$_{D}$C$_{D}$) \cite{li2019transferable} &A+S+P &47.8\\
        &VCL \cite{hou2020visual} &A+S &48.3\\
        &C-HOI \cite{zhou2020cascaded} & A+S & 48.3\\
        &DRG \cite{gao2020drg}  &A+S+L &51.0 \\
        &VSGNet \cite{ulutan2020vsgnet}  & A+S &51.7\\
        &PMFNet \cite{wan2019pose} &A+S+P & 52.0 \\
        &PD-Net \cite{zhong2020polysemy} &A+S+P+L &52.6\\
        &ACP \cite{kim2020detecting} &A+S+P+L &52.9\\
        &FCMNet  \cite{liuamplifying} &A+S+P+L&53.1\\
        &ConsNet \cite{liu2020consnet}  & A+S+L  &53.2\\
        &PD-Net \cite{zhong2021polysemy} &A+S+P+L &53.3\\
        \hline
        \hline
        \multirow{5}*{\rotatebox{90}{One-Stage}}
        &InteractNet \cite{Gkioxari2017Detecting} &A  &40.0 \\
        &UnionDet \cite{kim2020UnionDet} &A &47.5\\
        &IP-Net  \cite{wang2020learning} & A &51.0\\
        &PPDM-Hourglass$^{\circ}$ \cite{liao2020ppdm} &A &51.1\\
       &$\mathbf{GGNet}$-Hourglass     &A   &\textbf{54.7} \\
        \hline
        \end{tabular}}
         \label{VCOCO}
\end{table}

\begin{table*}[t]
\centering
\caption{Performance comparisons on HICO-DET. ``Backbone Sharing'' represents methods that use the same feature backbone for object detection and interaction detection.
$^{\diamond}$  denotes two-stage methods that first pre-train their object detectors on COCO, and then further fine-tune the object detectors on HICO-DET. $^{\ddag}$  denotes methods that use the same object detection results during inference.}
\resizebox{0.66\textwidth}{!}{
\begin{tabular}{c|c|c|c| ccc|ccc  }
\hline
&    &Backbone       & & \multicolumn{3}{c|}{DT Mode}  &\multicolumn{3}{c}{KO Mode}\\
& Methods  & Sharing & Feature  & Full & Rare & Non-Rare  & Full & Rare & Non-Rare \\
\hline
\hline
\multirow{12}*{\rotatebox{90}{Two-Stage}}
& No-Frills \cite{gupta2019no} &\checkmark &A+S+P    & 17.18  & 12.17     & 18.68     & -   & -   & -\\
& DRG \cite{gao2020drg} &- &A+S+L  &19.26 	 &17.74 	 &19.71 	 &23.40 	 &21.75 	 &23.89 \\
& Peyre \MakeLowercase{\textit{et al.}} \cite{peyre2019detecting} &\checkmark &A+S+L &19.40 &14.60 &20.90  & -   & -   & -  \\
& VCL \cite{hou2020visual} &- &A+S   &19.43 	&16.55 	&20.29 	&22.00 	&19.09 	&22.87 \\
& FCMNet  \cite{liuamplifying} &- &A+S+P+L   &20.41 	&17.34 	&21.56 	&22.04 	&18.97 	&23.12\\
& ACP \cite{kim2020detecting} &\checkmark &A+S+P+L  &20.59 	 &15.92 &21.98 & -   & -   & -\\
& PD-Net \cite{zhong2020polysemy} &\checkmark &A+S+P+L   &20.81 &15.90 &22.28 &24.78 &18.88 &26.54\\
& DJ-RN \cite{li2020detailed} &- &A+S+P+L   &21.34 &18.53 &22.18 &23.69 &20.64 &24.60 \\
& ConsNet \cite{liu2020consnet}  &\checkmark & A+S+L   &22.15 &17.12 &23.65  & -  &-   & - \\
& PD-Net \cite{zhong2021polysemy} &\checkmark &A+S+P+L   &22.37 &17.61 &23.79 &26.86 &21.70 &28.44\\
& VCL$^{\diamond}$\cite{hou2020visual} &- &A+S   &23.63 &17.21 &25.55 &25.98 &19.12 &28.03 \\
& DRG$^{\diamond}$$^{\ddag}$ \cite{gao2020drg} &-  &A+S+L   &24.53 &19.47 &26.04 &27.98 &23.11 &29.43\\
\hline
\hline
\multirow{8}*{\rotatebox{90}{One-Stage}}
&Shen \MakeLowercase{\textit{et al.}} \cite{Shen2018Scaling} &\checkmark &A+P   &6.46  & 4.24  & 7.12   & -   & -   & - \\
&InteractNet \cite{Gkioxari2017Detecting} &\checkmark &A   &9.94  & 7.16     & 10.77     & -   & -   & - \\
&UnionDet \cite{kim2020UnionDet} &\checkmark &A   &17.58  & 11.72  & 19.33  & 19.76  & 14.68  & 21.27 \\
&Our Baseline-Hourglass &\checkmark & A  & 21.43   &13.46 &23.81  &24.29 &16.40 &26.65 \\
&PPDM-Hourglass \cite{liao2020ppdm} &\checkmark & A & 21.73 &13.78 &24.10  & 24.58   & 16.65   & 26.84\\
& $\mathbf{GGNet}$-Hourglass      &\checkmark & A  &\textbf{23.47} & \textbf{16.48} &  \textbf{25.60} & \textbf{27.36} & \textbf{20.23} & \textbf{29.48} \\
&PPDM-Hourglass$^{\ddag}$  \cite{liao2020ppdm} &- & A  &26.50 &19.35 &28.63  & 29.24   & 22.18  & 31.34\\
& $\mathbf{GGNet}$-Hourglass$^{\ddag}$      &- & A   & \textbf{29.17} & \textbf{22.13} &  \textbf{30.84} & \textbf{33.50} & \textbf{26.67} & \textbf{34.89} \\
\hline
\end{tabular}}
\label{tab:hico}
\end{table*}

\subsection{Comparisons with State-of-the-Art Methods}
We compare the performance of GGNet with state-of-the-art methods on both V-COCO and HICO-DET.

As shown in Table \ref{VCOCO}, GGNet outperforms all state-of-the-art methods by significant margins on V-COCO. In particular, with the same backbone model, GGNet outperforms one of the most recent one-stage methods, i.e. PPDM \cite{liao2020ppdm} by  a large margin of 3.6\%  in terms of mAP. Moreover, although recent two-stage methods adopt various types of features to promote HOI detection performance, GGNet still outperforms all of these by at least 1.4\% in terms of mAP while utilizing appearance features only.

As shown in Table \ref{tab:hico}, on HICO-DET, GGNet still outperforms state-of-the-art methods by clear margins in different settings of ``Backbone Sharing''. In particular, GGNet outperforms PPDM  \cite{liao2020ppdm},  by  1.74\% (2.78\%), 2.70\% (3.58\%), and 1.50\% (2.64\%) mAP in DT (KO) mode for the full, rare and non-rare HOI categories, respectively.
Moreover, GGNet outperforms one of the best two-stage methods, i.e. PD-Net \cite{zhong2021polysemy}, by 1.10\% mAP in DT mode. Again, it is worth noting that PD-Net utilizes four types of features while GGNet employs the appearance feature only.

As Table \ref{tab:hico} shows, several recent two-stage methods  have adopted two separate backbones for object detection and interaction prediction. In particular, two approaches have fine-tuned both backbones on the HICO-DET databases, thereby achieving superior performance \cite{gao2020drg}, \cite{hou2020visual} (marked with $^{\diamond}$ in Table \ref{tab:hico}).  For its part, however, GGNet has a single shared backbone for object detection and interaction prediction. To facilitate fair comparison, we further test the performance of GGNet with the same object detection results provided by the object detector in DRG \cite{gao2020drg}; this setting is denoted as GGNet$^{\ddag}$. As shown in Table \ref{tab:hico}, GGNet$^{\ddag}$  outperforms DRG$^{\diamond}$$^{\ddag}$  \cite{gao2020drg} by a large margin of  4.64\% (5.52\%) mAP in DT (KO) mode.

The above comparisons also show that the adoption of a separate object detector can significantly promote GGNet’s performance. This may be because the object detection and interaction prediction tasks conflict with each other, as they require different features. It is also worth noting that adopting two backbones does not alter the one-stage nature of GGNet, as the object detection and interaction prediction tasks still run in parallel, regardless of whether or not they adopt the same backbone. The above experiments further justify the effectiveness of GGNet.
\subsection{Qualitative Visualization Results}
\begin{figure}[t]
	\begin{center}
		\includegraphics[width=0.48\textwidth]{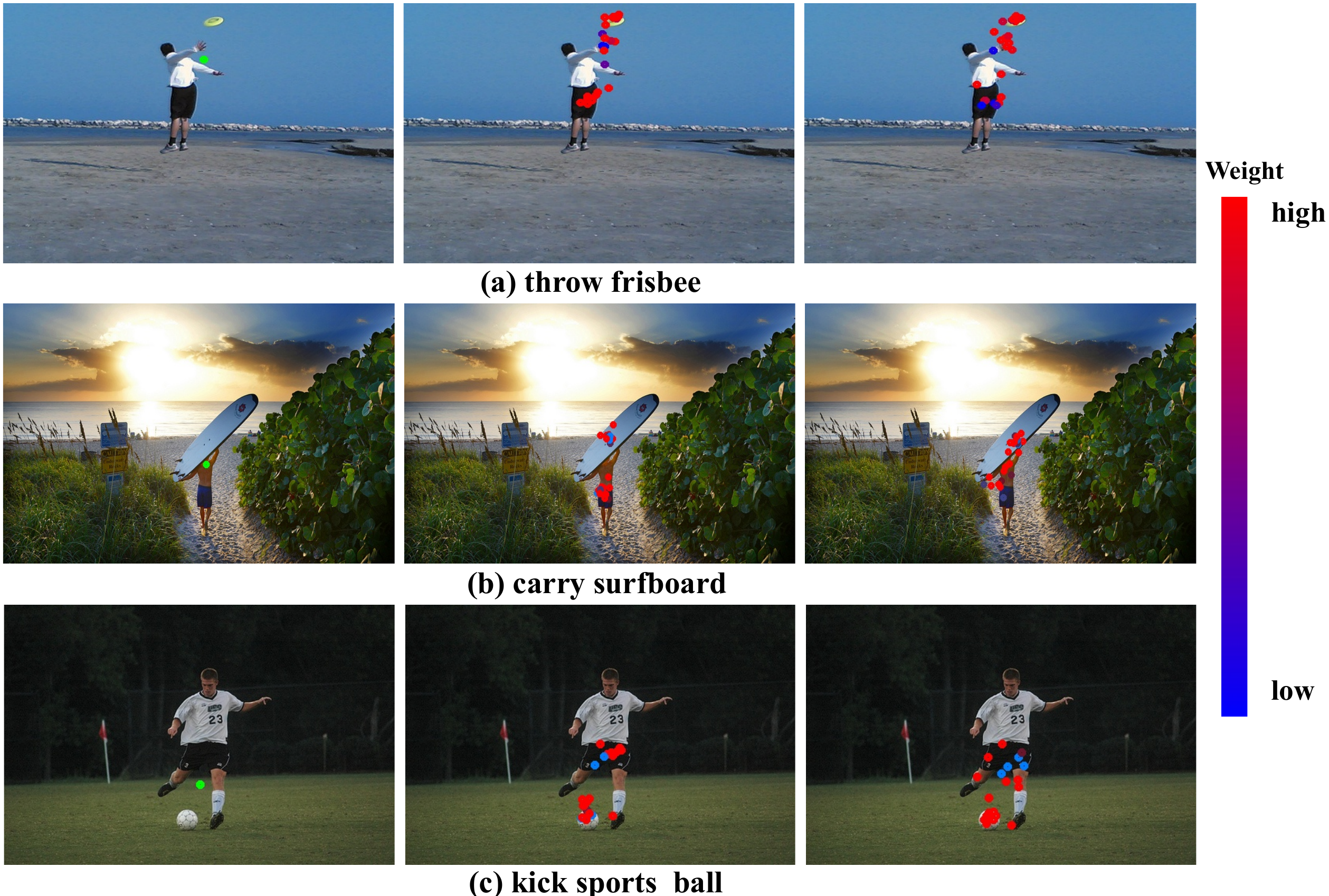}
	\end{center}
	\caption{Visualization of results in glance and gaze steps. The first column shows the detected interaction point in the Glance Step; the second and third columns visualize the adaptively sampled ActPoints by Gaze Step 1 and Gaze Step 2, respectively.  The color of each ActPoint reflects its weight, i.e. discriminative power. }
	\label{Figure:dcn_vis}
\end{figure}

Figure \ref{Figure:dcn_vis}  visualizes the interaction points and ActPoints predicted by GGNet. Here, three images are randomly chosen from V-COCO. We can observe that the interaction points, i.e. the green points in first column, are often located in the background area; therefore, their own features are vague to represent the interaction category. In comparison, ActPoints capture cues from discriminative object and human parts, both of which are important for interaction prediction. Moreover, the ActPoints refined by Gaze Step 2 are usually located at more important object and human parts than the ActPoints sampled by Gaze Step 1. In the supplementary material, we also present qualitative comparisons between GGNet and PPDM in terms of HOI detection results.
\section{Conclusion}
Existing one-stage HOI detection methods  typically  utilize the features of predefined interaction areas for interaction prediction, while these artificially defined areas are usually vague to represent interactions. In this paper, we propose a novel one-stage network, namely GGNet, which adaptively samples a set of action-aware points (ActPoints) via glance and gaze steps. We also propose an action-aware point matching approach that robustly matches target human and objects for each detected interaction. Moreover, a novel hard negative attentive loss is devised to improve the optimization of GGNet. Extensive experiments results
show that GGNet outperforms start-of-the-arts on both V-COCO and HICO-DET datasets.

\noindent{\bf Acknowledgement}
This work was supported in part by the National Natural Science Foundation of China under Grant 62076101, 61702193, and U1801262, in part by the Program for Guangdong Introducing Innovative and Entrepreneurial Teams under Grant 2017ZT07X183.


\clearpage

\begin{appendices}

This supplementary material includes five sections.
Section \ref{sec:overview} illustrates the structure of GGNet in the inference stage.
Section \ref{sec:HNA} conducts ablation study on the value of hyper-parameter $\beta$.
Section \ref{sec:lambda} carries out the  sensitivity analysis for $\lambda_{1}$.
Section \ref{sec:FG} shows the structure of ``Variants of Feature Aggregation'' in Section 5.2 of the main paper.
Section \ref{sec:vis} visualizes the HOI detection results of GGNet and PPDM; some failure cases of GGNet  for HOI detection are also presented here.

\section{Structure of GGNet in the Inference Stage}
\label{sec:overview}
Figure \ref{Figure:overview} illustrates the structure of GGNet in the inference stage.
During inference, the glance step and the first gaze step are only utilized to infer ActPoints; therefore, some layers in these two steps are removed.

\section{Ablation Study on the Value of $\beta$}
\label{sec:HNA}
Experiments are conducted on the V-COCO database.
The experimental results are summarized in Table \ref{table:hn ablation1}.
We can observe that the HNA loss achieves the best performance when $\beta$ is set to 7.
\begin{table}[h]
	\centering
	\caption{Ablation study on the value of $\beta$.}
	\vspace{1mm}
	\resizebox{0.25\textwidth}{!}{
		\setlength\tabcolsep{2pt}
		\renewcommand\arraystretch{1.1}
		\begin{tabular}{c|c|c|c|c}
			\hline
		    \text{$\beta$} & 5 & 6 & 7 & 8 \\ \hline
		    $\text{mAP}_{role}$ & 54.33 & 54.28 & \textbf{54.72} & 54.45 \\ \hline
		\end{tabular}
	}
	\label{table:hn ablation1}
\end{table}

\section{Sensitivity analysis for $\lambda_{1}$}
\label{sec:lambda}
Experiments are conducted on the V-COCO database.
The experimental results are listed in Table \ref{table:ablation2}.
We can observe that the GGNet achieves the best performance when $\lambda_{1}$ is set to 0.1.
\begin{table}[h]
	\centering
	\caption{Sensitivity analysis for $\lambda_{1}$.}
	\vspace{1mm}
	\resizebox{0.25\textwidth}{!}{
		\setlength\tabcolsep{2pt}
		\renewcommand\arraystretch{1.1}
		\begin{tabular}{c|c|c|c}
			\hline
		    \text{$\lambda_{1}$} & 0.1 & 0.5 & 1  \\ \hline
		    $\text{mAP}_{role}$ &  \textbf{54.72} & 54.01 & 53.28\\ \hline
		\end{tabular}
	}
	\label{table:ablation2}
\end{table}

\section{Model Structure of Variants for Feature Aggregation}
\label{sec:FG}
In this section, we show the structure of ``Variants of Feature Aggregation'' in Table 3 of the main paper.
All methods in Table 3 share the same structure of the human-object pair matching module as the baseline model.
Besides, they all adopt the ordinary $V$-dimensional element-wise focal loss [10] for optimization.

\begin{figure*}[!t]
	\begin{center}
		\includegraphics[width=0.84\textwidth]{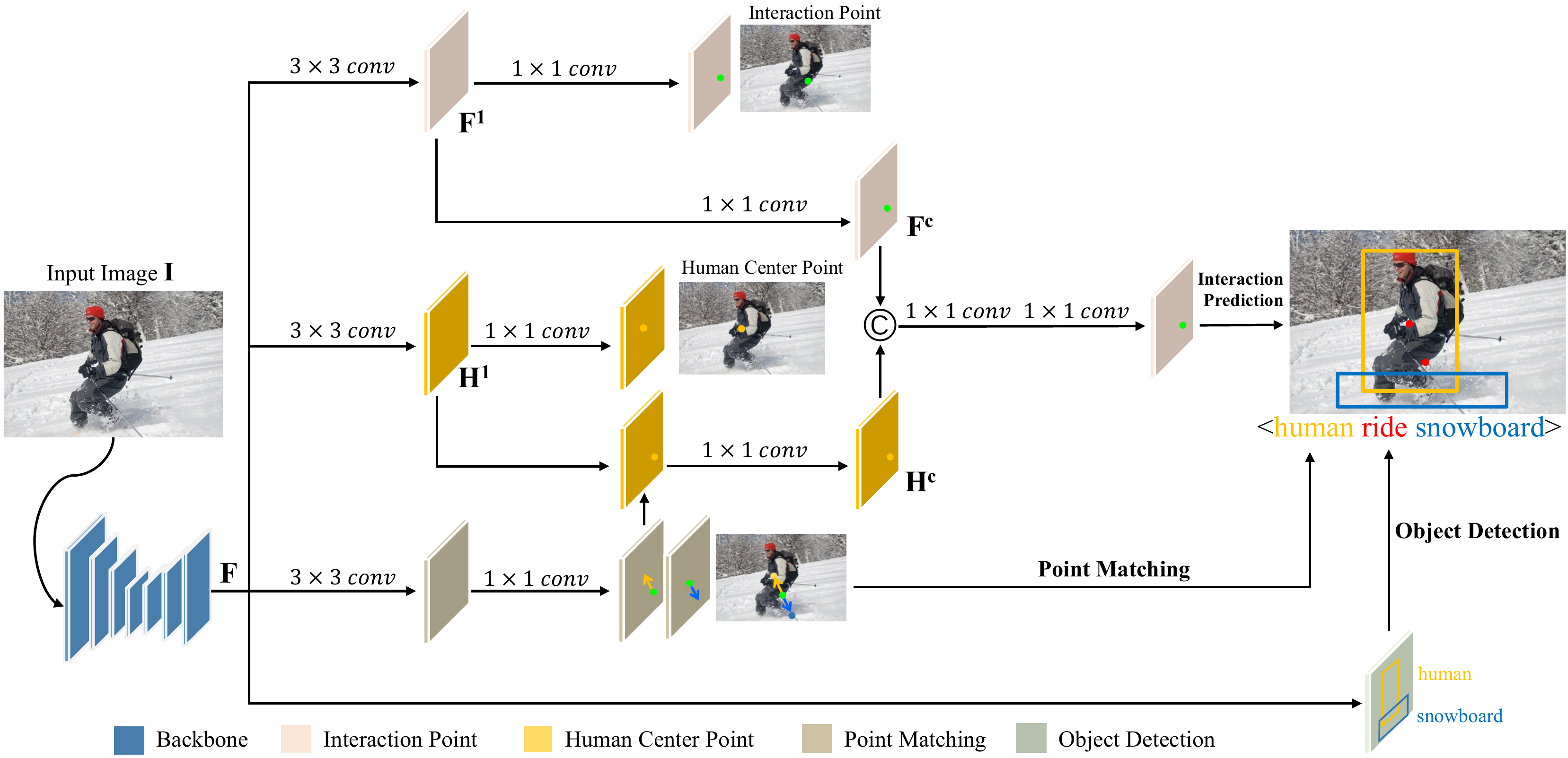}
	\end{center}
	\caption{ Overview of the model ``I + H'' in the training stage.
	The model is composed of four branches, namely the interaction point detection branch, the human branch, the point matching branch, and the object detection branch.
	The four branches run in parallel.
	Features of each interaction point and those of the corresponding human center point are concatenated for interaction prediction.
	$\copyright$ denotes the concatenation operation in the channel dimension.
	}
	\label{Figure:overview_HO}
\end{figure*}

\noindent{\bf The structure of Model ``I + H'' (``I + O'')}.
The model ``I + H'' is illustrated in Figure \ref{Figure:overview_HO}.
To obtain the human feature for each human-object pair, we first attach one human (H) branch on the backbone model.
The H branch runs in parallel with the interaction point detection (I) branch.
The H branch is realized using a 3 $\times$ 3 Conv layer with ReLU, followed by a 1 $\times$ 1 Conv layer and a sigmoid layer.
To enable the feature maps $\mathbf{H}^{\mathbf{1}}$ to be action-aware, we apply a $V$-dimensional element-wise focal loss to the H branch as supervision.

Next, we utilize the offset predicted by the human-object pair matching module, i.e. the point matching branch in  Figure \ref{Figure:overview_HO}, to predict the  human center point for each interaction point in $\mathbf{F}^{\mathbf{1}}$.
Features of the human center point are extracted on $\mathbf{H}^{\mathbf{1}}$ using the bilinear sampling [26] and are further concatenated with the features of the interaction point.
The concatenated features are processed by two successive 1 $\times$ 1 Conv layers for interaction prediction.



The model ``I + O'' can be constructed in a similar manner by replacing the above H branch  with an object (O) branch.
Features of the object center are utilized to augment the features of the corresponding interaction point.

\noindent{\bf The Structure of  Model  ``I + H + O''}.
This model can be constructed by adding both the H and O branches.
The features of human center, object center, and the interaction point are concatenated for interaction prediction.

\section{Qualitative Visualization Results}
\label{sec:vis}
\begin{figure*}[t]
	\begin{center}
		\includegraphics[width=0.98\textwidth]{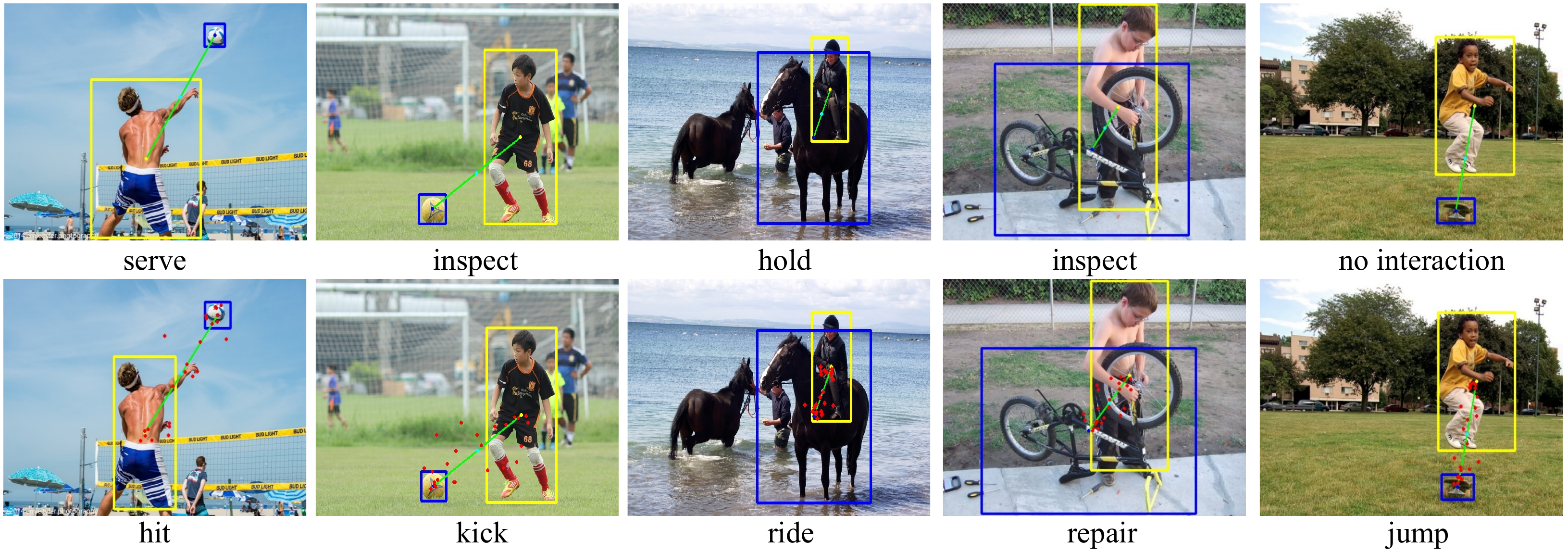}
	\end{center}
	\caption{Qualitative comparisons between GGNet and PPDM on HICO-DET.
	The first and second rows show the predictions by PPDM and GGNet respectively.
	Cyan denotes the interaction points and  red stands for ActPoints. Moreover, the human and objects are represented using yellow and blue, respectively.
	If a person has interaction with an object, they are linked by a green line. We show the top-1 triplet  according to the prediction confidence per image.
	}
	\label{Figure:cmp_ppdm_ggnet}
\end{figure*}

\begin{figure*}[!t]
	\begin{center}
		\includegraphics[width=0.98\textwidth]{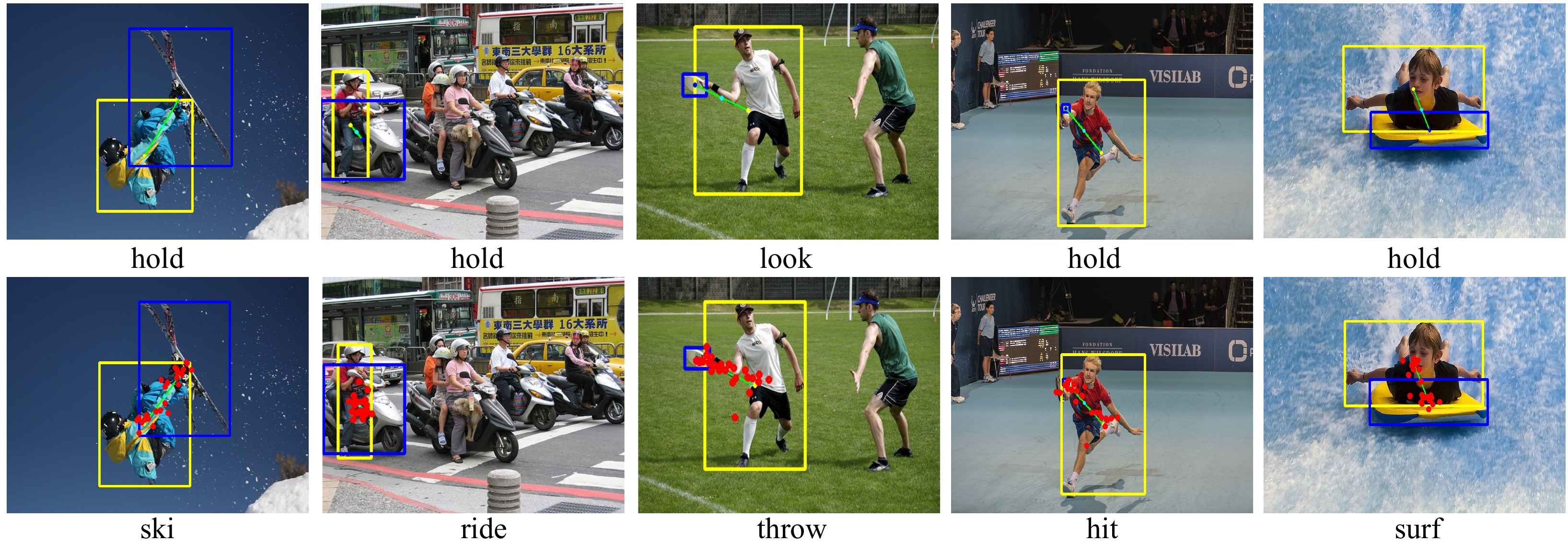}
	\end{center}
	\caption{Qualitative comparisons between GGNet and PPDM on V-COCO.}
	\label{Figure:cmp_ppdm_ggnet2}
\end{figure*}

Figure \ref{Figure:cmp_ppdm_ggnet} presents the qualitative comparisons between GGNet and PPDM [10] in terms of HOI detection results on HICO-DET.
We can observe that PPDM fails to predict interaction categories for some images.
This is because the interaction points often locate at the background area or unimportant human body area; therefore, their features are  ambiguous in semantics for interaction prediction.
In comparison, GGNet infers the interaction categories accurately, as the discriminative interaction areas can be captured by our proposed glance-and-gaze strategy.
Qualitative comparisons on  V-COCO are shown in Figure \ref{Figure:cmp_ppdm_ggnet2}.

\begin{figure*}[!t]
	\begin{center}
		\includegraphics[width=0.98\textwidth]{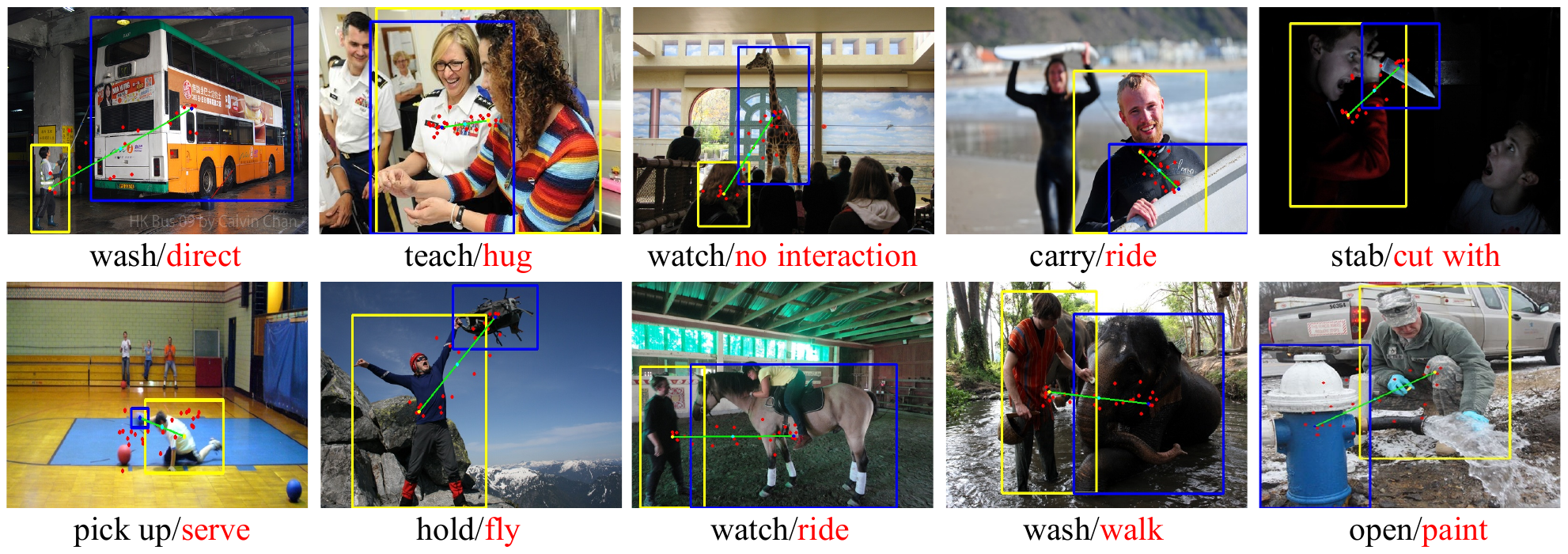}
	\end{center}
	\caption{Failure cases of GGNet for HOI detection on HICO-DET. The ground-truth interaction and the  predicted one are typed in black and red, respectively. }
	\label{Figure:failure_case}
\end{figure*}

We also present some failure cases of GGNet  in terms of HOI detection in Figure \ref{Figure:failure_case}.

\begin{figure*}[!t]
	\begin{center}
		\includegraphics[width=0.84\textwidth]{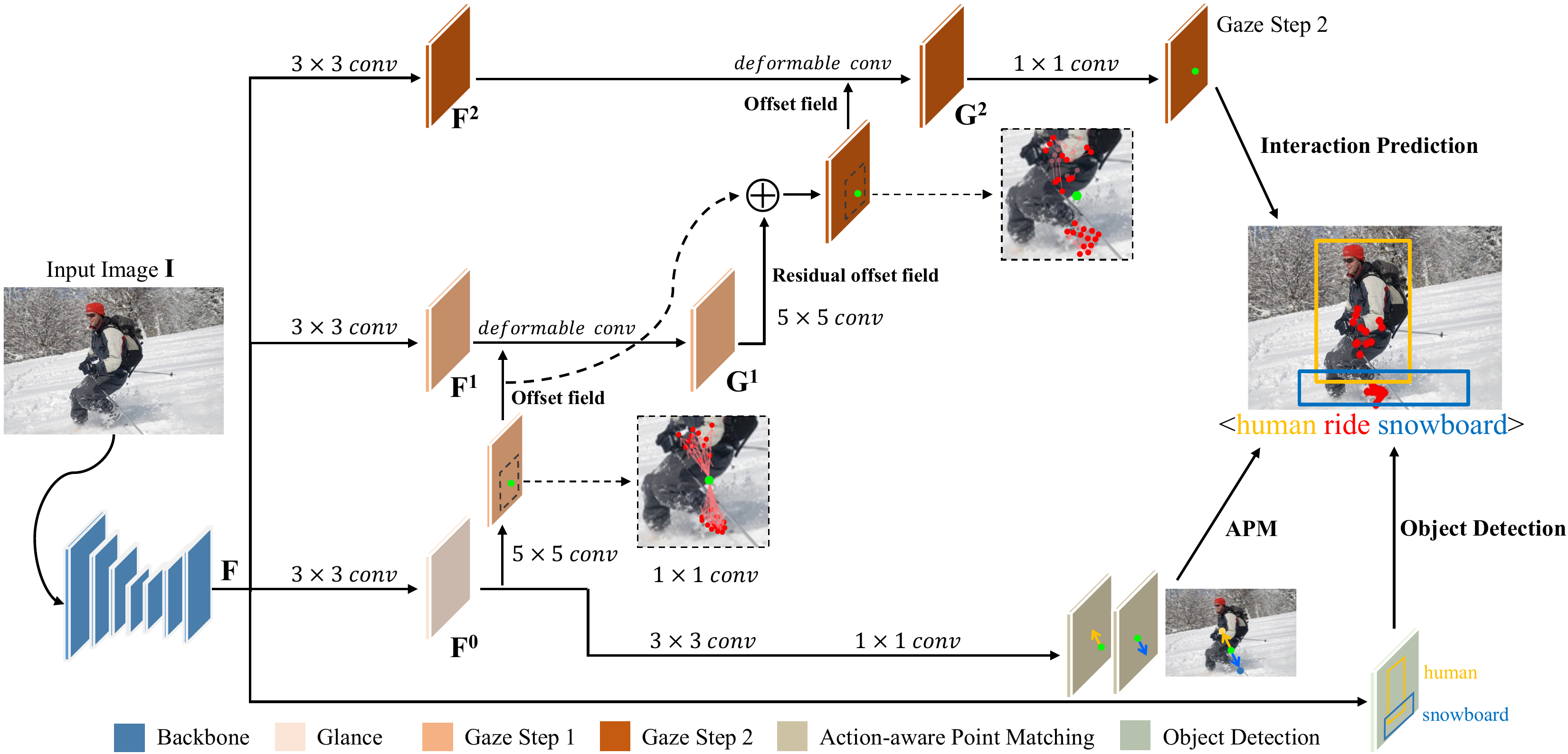}
	\end{center}
	\caption{Overview of GGNet in the inference stage.
	The Glance step and Gaze Step 1 are only used to infer the ActPoints,
	and the irrelevant layers are discarded.}
	\label{Figure:overview}
\end{figure*}

\end{appendices}

\end{document}